\documentclass[9.5pt, journal]{IEEEtran}
\ifCLASSINFOpdf
\else
\fi
\usepackage{cite}
\usepackage{graphicx}
\usepackage{mathtools}
\usepackage{textcomp}
\usepackage{multirow}
\usepackage{algpseudocode}
\usepackage{amssymb}
\usepackage[ruled]{algorithm2e}
\usepackage{xcolor}
\usepackage{mathptmx}
\usepackage[11pt]{moresize}
\begin{document}

\title{ A Distributed Acoustic Sensor System for Intelligent Transportation using Deep Learning}

 \author{Chia-Yen Chiang, Mona Jaber,~\IEEEmembership{Senior Member,~IEEE,} and~Peter Hayward
 \vspace{-1cm}        
 \thanks{C. Chiang and M. Jaber are with the School
 of Electronic Engineering and Computer Science, Queen Mary University of London, London E1 4FZ, UK e-mail: c.chiang@qmul.ac.uk and m.jaber@qmul.ac.uk.}
 \thanks{P. Hayward is with Fotech Group, Church Crookham GU52 0RD, UK e-mail: peter.hayward@fotechsolutions.com}}


\maketitle

\begin{abstract}
Intelligent transport systems (ITS) are pivotal in the development of sustainable and green urban living. ITS is data-driven and enabled by the profusion of sensors ranging from pneumatic tubes to smart cameras. This work explores a novel data source based on optical fibre-based distributed acoustic sensors (DAS) for traffic analysis. Detecting the type of vehicle and estimating the occupancy of vehicles are prime concerns in ITS. The first is motivated by the need for tracking, controlling, and forecasting traffic flow. The second targets the regulation of high occupancy vehicle lanes in an attempt to reduce emissions and congestion. These tasks are often conducted by individuals inspecting vehicles or through the use of emerging computer vision technologies. The former is not scale-able nor efficient whereas the latter is intrusive to passengers' privacy. To this end, we propose a deep learning technique to analyse DAS signals to address this challenge through continuous sensing and without exposing personal information. We propose a deep learning method for processing DAS signals and achieve 92\% vehicle classification accuracy and 92-97\% in occupancy detection based on DAS data collected under controlled conditions.
\end{abstract}

\IEEEpeerreviewmaketitle
\vspace{-0.5cm} 
\section{Introduction}
\IEEEPARstart{I}{ntelligent} transportation systems (ITS) analyse traffic data to automate traffic management decisions that would result in environmental-friendly and efficient traffic-flow. One ITS interest is to encourage vehicles with multiple occupants by allowing these to reduce their journey times relative to single-occupant vehicles, particularly when the general purpose lanes are congested. To this end, High Occupancy Vehicle (HOV) lanes are designed to prioritise HOVs and buses. The expected outcome is the reduction of vehicles on the road network which would positively impact the overall congestion, CO2 emissions, and fuel consumption. Various trials in Europe and the USA resulted in time reduction of trips on HOV corridors by 4\% to 30\%\cite{leeds}. It has also been reported that willingness to share cars and to use buses increases after the opening of an HOV lane. Current methods used to enforce HOV lanes either require the physical presence of police officers to visually monitor the number of passengers in cars or encourage commuters to report offenders of HOV lanes. Recently, camera-based technologies such as~\cite{thermalimages01} and~\cite{HOVHOT} have been proposed for automating the passenger counting task, however these are costly and intrusive.

Automatically calculating passenger numbers has other uses in ITS applications such adapting the reach and operation of public transportation facilities with new infrastructure~\cite{estimatedemand} or estimating~\cite{waitingtime} and optimising\cite{decreasewaiting} traffic waiting time by developing adaptive traffic light system. The lack of cost-effective methods to detect the vehicle occupancy limits these applications and the potential of ITS to enable sustainable commuting as an optimised mix of public and private transport means.
For instance, a common way to restrict low-occupancy vehicles in Taiwan is to hire police to do raw-eye checking on highway entrance. It is usually dangerous, highly-repetitive, time-consuming and expensive. In South Korea and The USA, traffic monitoring staff check occupancy through camera but still cannot prevent such labouring task. To predict occupancy~\cite{infraredimage}, two infrared camera were assigned in a fixed location to automatically capture front-rear seat occupation information which is sent to an object detection model for car occupancy prediction. This method is attractive for being privacy-preserving as has been demonstrated in~\cite{infraredimage} and~\cite{JAIN}. However, such technology is limited by specific location where the sensor is situated. The hardware cost cannot be negligible if the authority uses it in the public. To boost accuracy, the model is more computationally expensive compared to usual machine learning classifier, since it is trained on images with multiple complex information.  

The other essential ITS objective with a rising interest is vehicle  classification and flow. The vehicles' characteristics of interest to ITS are multi-fold and include the size and the engine type, as each would have different impact on urban planning and environmental pollution. In recent years, numerous traffic detectors have been deployed to monitor traffic flow such as remote traffic microwave sensors~\cite{RTMS}, magnetic sensors~\cite{magnetic}, in addition to more traditional sensors such as pneumatic road tubes~\cite{pneumatic}. However, the data generated by these sensors is mostly limited to counting of vehicles and optimising the flow accordingly by controlling traffic lights, for instance. However, the information obtained by these sensors in not sufficient to solve emerging ITS problems such as the specific flows of vehicles with combustion engine or batteries. 

More recently, video sensors have been deployed for traffic flow monitoring with the ability to classify the type of passing vehicles and to track their movements~\cite{TrafficForecast}. For instance, authors in~\cite{ITS} employ advanced deep learning techniques in the analysis of video footage to detect and track vehicle movement. It is, however, extremely challenging and costly to cover and analyse kilometers of road networks with video cameras and the limitation due to blind areas would be inevitable.
\section{Contributions}
In this work we examine how Distributed Acoustic Sensor (DAS) systems could be used as an alternative source for classifying vehicle types and occupancy. DAS reuses underground fiber optic cables as a distributed strain sensing system where the strain is caused by moving objects above ground. Research on DAS was initially targeted toward the monitoring of oil and gas pipelines, peripheral safety, structural health, and submarine power cables.  
There are several advantages for using DAS systems in ITS. Firstly, fibre optic lines are often available and can be readily reused as DAS. Second, DAS technology is not affected by adverse weather conditions nor light intensity. Thirdly, the movement signature detected in DAS does not contain private information such as people's faces, clothing, or identity. Last, DAS offers an uninterrupted continuous sensor system over up to 40 to 50 kilometres without obstructions such as those impacting computer vision based solutions~\cite{book_DOFS}.

We present the first work that leverages DAS systems as a solution for ITS. Due the complexity and massive size of the DAS data, machine learning and deep learning in particular are suggested to model the key features that help answer ITS-related questions. To this end, we propose to employ convolution neural networks (CNN) trained with both one-dimentional (1D) DAS time series and two-dimentional (2D) spacio-temporal DAS segments. The results for both occupancy (up to 97\% accuracy) and vehicle size (up to 92\% accuracy) classification reinforce our thesis that the DAS signal contains key information for ITS applications that can be mined using CNN. In order to motivate more research in this area, we have setup a github repository for this project that will become accessible upon the publication of the article (Link: https://github.com/Chiayen0503/DAS-signal-preprocessing-and-classification).

\section{Related Research}\label{sec:lit}
To estimate a vehicle's occupancy rate, the actual passenger number and the nominal capacity of the vehicle are needed. There are few conventional sensor-based technologies used in passenger flow estimation: (1) Closed-circuit television (CCTV)/{Thermal images} (2) Infrared sensor (3) Impulse Radio Ultra-104 Wideband (IR-UWB) radar sensors. CCTV is commonly used in monitoring real-time visual information in a given area. In~\cite{bus1}, two CCTV cameras were deployed on board of a bus to capture the full picture of total passenger heads. A \textit{You Only Look Once} (Yolo3) and a Convolutional Autoencoder (CAE) were trained to detect passenger heads in clear and blurred areas affected by occlusion, respectively. However, the CAE model wasn't sophisticated to deal with light exposure in blurred and obstructed areas. This is one of the common issues of computer vision; model prediction performance is heavily influenced by extreme lightness, wide-angle distortion and blind spot. In addition, compared to (2) and (3), CCTV may cause privacy breach concerns. According to the General Data Protection Regulation (GDPR)~\cite{GDPR01}, the public should be informed of what data is being recorded, its use, and how long the footage is stored. Therefore, it is difficult to widely implement such technology for monitoring occupancy rate in private vehicles. {Although this concern can be eliminated by changing the data source to thermal images~\cite{thermalimages01}, the solution comes at high cost and the blind-spot problem persists where thermal cameras cannot see through surface with high reflective index such as glass or wall~\cite{thermalimages03}.}  

Infrared light based methods count passenger number by fitting a {Radial Basis Function neural network (RBF)~\cite{RBF}} on pulse data collected by the infrared light sensor~\cite{infrared01}. Alternative methods, such as~\cite{infrared02}, use K-nearest neighbours to classify motion direction based on the distances measured by three infrared sensors. The former method suffers from decrease in detection accuracy when more passengers pass simultaneously. The latter method only works when a single person passes through the sensors. 

The limitations in~\cite{infrared01} and~\cite{RBF} can be circumvented by using IR-UWB radar sensors. In~\cite{iruwb2}, for instance, the model error rate for detecting multiple people passing simultaneously through the subway entrance is 10\%. The number of detected passengers could be up to 9000 or more or as small as single digit in a day. The deployed module for human detection contains two sensors which receive two signals respectively. The signals were then calculated for their maximum correlation which served as an indicator of human detection and direction detection. The application was cost-effective and easily setup but with the same location limitation like camera sensor.

As mentioned in the previous section, DAS has been used for the detection and classification of acoustic events, but not for the ITS purposes covered in this work. In line with acoustic signals classification, manual feature extraction techniques were first adopted in DAS event classification. Such techniques include wavelet packet transform~\cite{VehClass1,Pattern2}, spectral substitution~\cite{Pattern3}, Mel-spectrograms~\cite{EventRec4}, and empirical mode decomposition~\cite{EventRec5}. The classification task is then conducted using conventional classifiers such as support vector machines (SVM such as~\cite{VehClass1}) and relevant vector machines (e.g.,~\cite{Pattern2}). In order to improve the classification results, authors in~\cite{EventRec5} use XGboost, an ensemble algorithm, and authors in~\cite{Pattern3} and~\cite{EventRec4} use CNN, achieving higher success rate than conventional classifiers. Instead of manual feature extraction, authors in~\cite{1DCNN} employ a 1D-CNN directly to the raw signal followed by an SVM-based classifier and outperform previous works.

This work is the first to propose a deep learning approach for DAS feature extraction that can be leveraged in ITS for detecting the size and occupancy of vehicles. Advances in intelligent traffic flow such as Cooperative Adaptive Cruise Control (CACC) have raised the question of re-purposing existing road networks, and HOV lanes in particular, to reduce congestion~\cite{CACC}. In this context, it is necessary to distinguish the sizes and occupancy of vehicles moving within the road network. This work offers a data-driven, scale-able, and privacy-preserving solution that is resilient to adverse weather and visibility conditions. The proposed solution reveals size-specific traffic flows and related occupancy, thus, informing ITS emerging objectives.

\section{DAS: System and Data}\label{sec:DAS}
In this section we present a brief overview of DAS systems and of the real DAS data used in this research. 

\subsection{DAS system}\label{sec:DASsys}
DAS system is an opto-eletronic device sensitive to strain distribution which measures the variation of force in different parts of fibre (up to a total fibre length of 40-50km). DAS is based on Optical Time Domain Reflectometry (OTDR) where a pulse of coherent light is periodically injected in the fibre. A fraction of this pulse gets reflected through back-scattering (Rayleigh) mechanism and is captured by a photodetector. This phenomena creates a continuous time-series of back-scatter intensity which is commonly called "fibre shot", where a shot corresponds to a time unit that is proportional to the time that a pulse takes to travel along fibre. In OTDR, back-scatter intensity decreases exponentially along the fibre distance except when a sudden variation on fibre-reflectivity coefficient occurs due to a faulty splice and fibre break. In contrast, in DAS systems, the intensity is modelled as a random function of fibre position and the cumulative phase of the interference of back-scattered signal and interrogating pulse. A "resolution cell" refers to the fibre section that gives rise to the back-scatter interference. Although interference phase is inherently random because of random molecular structure of the fibre glass, it remains constant if the state of fibre is unchanged. On the other hand, if the dynamic strain varies, the back-scatter phase and its intensity will vary accordingly. Therefore, many different measurements are used to evaluate the variation of back-scatter at given fibre distance to draw insights of how the strain evolves dynamically in a given fibre position.

The width of probing pulse decides the size of resolution cell which determines the DAS spatial resolution. A pulse with 100~ns corresponds to 10~m spatial resolution. Because the number of sensing unit is decided by spatial sample period, the smaller the period the more likely a pulse with the same type of light can be captured by multiple sensing units at once. On the other hand, temporal sampling resolution (TSR) is given by interrogating pulse repetition frequency which is limited by fibre length because it becomes harder to complete a cycle of pulse when the pulse's travelling length is increased. TSR refers to how often data of the same fibre position is collected and has a trade-off relation with the spatial resolution. Hence, the output of DAS is a collection of digitised fibre shots retrieved by given TSR. DAS can be seen as a two dimensional spatio-temporal array of scalar measurements.
\subsection{DAS data}\label{sec:dataA} 
In a controlled field trial, DAS data was collected on a 4.8km road stretch equipped with a DAS system, as described in Section~\ref{sec:DASsys}. The fibre was buried 20cm in a micro-trench and Fotech's Helios \footnote{https://www.fotech.com/technology/helios-das/} was used to capture the DAS signals. Two use cases were considered: (a) driving the same car with different occupancy (see Figure~\ref{fig1}), (b) driving five different vehicles in a pre-defined queue with controlled speed (see Figure~\ref{fig:2Dplot}).
In Figures~\ref{fig1}) and~\ref{fig:2Dplot}, the x-axis indicates the time in fibre \textit{shots}, where one shot $s$ is equivalent to $1/1000.04$ seconds. The y-axis shows the position along the fibre in \textit{bins} where one bin $b$ is equivalent to $0.68$ metres. The color intensity of each pixel, at bin $b$ and shot $s$, shows the strength of displacement in radians $\rho(b,s)$. There are ten dominant lines composed of clear white-black pixels across-multiple-bins. These are car signals that have generated a linear disturbance as a result of fixed speed and are highlighted in red in Figure~\ref{fig1} and \ref{fig:2Dplot}. During data collection, one side of the road was blocked to reduce other vehicles driving and interfering with the controlled DAS data collection. The data collected for each use case is detailed in the following sections.

\subsubsection{Renault Clio at 60km/h with different occupancy (RC-60-Mix)}
\label{sec:MainData}
\begin{figure}[h!]
\centering
\includegraphics[width=0.9\columnwidth]{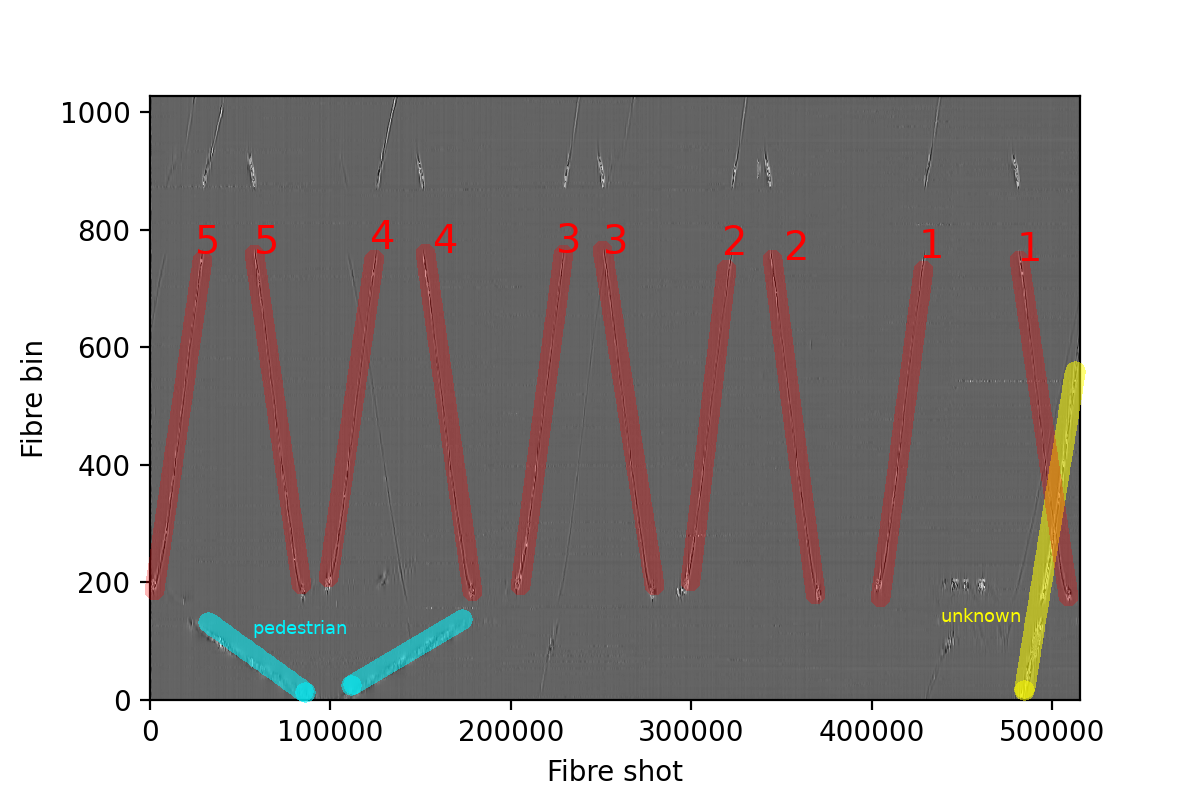} \vspace{-0.5cm}
\caption{Ten highlighted lines show car moving at a constant speed of 60~km/h. Each line is labelled with the number of passengers (RC-Mix). 
\label{fig1}}
\end{figure}

In the experiment, a Renault Clio \textbf{RC} (maximum capacity is five passengers) was driven at a fixed speed of 60~km/h in two different directions of the road stretch: east-to-west and west-to-east. Each trip was repeated in a predefined order with different number of passengers: from five to one, as highlighted in red in Figure~\ref{fig1}. Three other tracks/lines can be seen; these represent the movement of other objects outside the controlled experiment. The blue lines between $0\leq b\leq 200$ indicate an approximate speed of 5km/h, hence may be the speed of a person speed walking in both directions. The yellow line between $0\leq b\leq 500$ on the right side of Figure~\ref{fig1} indicates the motion of a fast object at an approximate speed of 60~km/h, which may be a vehicle outside the controlled experiment.           
\vspace{-0.5cm}
\subsubsection{Renault Clio at 60km/h with five passengers (RC-60-5p)}
\label{sec:Ind5P}
Similar to the experiment described in Section~\ref{sec:MainData} and shown in Figure~\ref{fig1}, the same Renault Clio is driven in both directions at the same speed of 60km/h with five passengers at all times. This dataset is used later as an independent data to test for model overfitting (see Table~\ref{tab:classificationdata}).
\subsubsection{Five vehicles at different speeds  (AllCars-1p)}\label{sec:sizeData}
\begin{figure}[!h]
\centering
  \includegraphics[width=0.9\columnwidth]{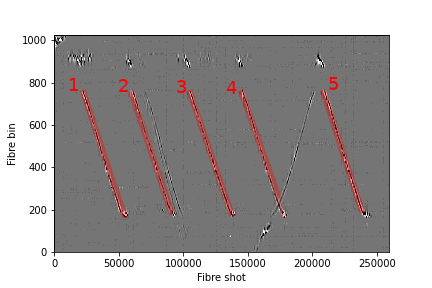}
  \caption{Section of DAS collected data for 50 km/h experiment showing the five target vehicles in addition to two unwanted ones (taken from AllCars-1p).}\label{fig:2Dplot}
\end{figure}

In this case, five different vehicles were driven (no other passengers but the driver) at different fixed speeds (30~km/h, 40~km/h, 50~km/h, 60~km/h, 70~km/h). 
Figure~\ref{fig:2Dplot} shows a section of the collected data for the 50~km/h speed dataset. The dominant five lines highlighted in the figure, represent the track of the five vehicles moving in a queue at the same speed and respecting the same inter-distance. Two other lines can be seen in Figure~\ref{fig:2Dplot}. The first represents another vehicle outside the controlled experiment that is moving in the same direction and is positioned between Car~2 (RC) and Car~3. The other one represents another external vehicle that is moving in the opposite direction and overlaps with both Car~4 and Car~5. Since this work is concerned with detecting the size of a vehicle and its corresponding occupancy, we group the five vehicles into two groups. The first group, \textit{Large}, is created by combining data from Car~1 (Sports utility vehicle) and Car~5 (Light commercial vehicle) as they are the top two largest vehicle types. The second group, \textit{Small}, includes Car~2 (RC), Car~3 (compact) and Car~4 (multi purpose vehicle).

\section{Problem formulation}\label{sec:problem}
In this section, we formulate the problem of estimating the selected car's occupancy and size, based on the imprinted DAS signature. Consider a raw DAS dataset of $B$ fibre bins collected over a duration of $M$ fibre shots. Each fibre bin $1\leq b\leq B$ represents a virtual acoustic sensor that measures the displacement of the optical signal in Radian at any time $1\leq s\leq M$. Let $\rho \in\mathbb{R}^{B \times M}$ be this complete raw dataset over the entire length of the fibre $B$ and over the full period of recording $M$. 

\subsection{Occupancy detection (5-way)}\label{Sec:Occ}
Let $\mathbf{x}^{\varphi,\Psi}_b \in \mathbb{R}^{d^{\varphi}_b}$ be a data sample of index $1\leq b\leq B$ that is collected from bin/sensor $b$ during the duration $d^{\varphi}_b<M$, where $1\leq\varphi\leq 5$ represents the number of passengers in a given car ($\varphi=0$ refers to an unwanted event, i.e., noise or unwanted vehicle) and $\Psi=[east,west]$ indicates the direction of travel. Based on the controlled experiment setup described in Section~\ref{sec:dataA}, $\exists!~d^{\varphi}_b>0$ for each value of $1\leq \varphi\leq 5$ at each bin $b$ in each direction $\Psi$, where
$d^{\varphi,\Psi}_b=s^{\varphi,\Psi}_{b,e}-s^{\varphi,\Psi}_{b,s}$ for all bins $1\leq b\leq B$, $s^{\varphi,\Psi}_{b,s}$ is the starting shot of event $\varphi$ at bin $b$ in direction $\Psi$, and $s^{\varphi,\Psi}_{b,e}$ is the end shot. Let $d^{\varphi}=\max (d^{\varphi,\Psi}_b)$ be the largest duration for an event $1\leq \varphi\leq 5$ for all values of $1\leq b\leq B$. Following the order of the occupancy depicted in Figure~\ref{fig1}, it can be said that for any bin $b$, $s^{5,\Psi}_{b,s}<s^{4,\Psi}_{b,e}<s^{3,\Psi}_{b,s}<s^{2,\Psi}_{b,e}<s^{1,\Psi}_{b,s}<s^{1,\Psi}_{b,e}$ and all other events represent noise (i.e., $\varphi=0$). Moreover, for every bin $b$, shot $s$, and any $1\leq \varphi\leq 5$, $s^{\varphi,east}_{b,s}<s^{\varphi,west}_{b,s}$.

We can then express the set of all data samples $ \mathbf{x}^{\varphi}_b$ in the dataset of passenger number $\varphi$ as $X^{\varphi} \in \mathbb{R}^{B \times d^{\varphi}}$ as the sequence of $B$ data samples describing event $\varphi$ that are collected by $B$ sensors for a duration of $d^{\varphi}$ seconds. The starting shot and end shot of each data sample in $X^{\varphi}$ differ, and they each span a different duration; in this work $d^{\varphi}$ is the largest duration of an event $\varphi$ within a sequence $X^{\varphi}$. Similarly, we can express $\mathbf{X} \in \mathbb{R}^{B \times d^{\varphi} \times 5}$ as the union of all such sequences $X^{\varphi}$ in the data set for all $1\leq\varphi\leq 5$ where each data sample $ \mathbf{x}_b$ in $X^{\varphi}$ (we drop the superscript $\varphi$ since all samples in $X^{\varphi}$ have the same $\varphi$) is associated with a label $y_b=\varphi$ and $b=1\dots B$.

The problem can thus be formulated as a \textit{regression} predictive modelling; a process of predicting passenger number that is categorised into five labels (5-way) that correspond to the number of passengers $\varphi$ by approximating a mapping function from input data samples in $\mathbf{X} = \{X^1, \cdots, X^5\}$ into discrete output labels $Y$. To this end, we reorganise the set of input data samples $\mathbf{X}$ into a training set $\mathcal{D}^t$ and a testing set $\mathcal{D}^v$. $\mathcal{D}^t \in \mathbb{R}^{D^t \times d^m}$ is a two dimensional matrix where $D^t$ is a design parameter that indicates the number of samples included in the training set and $d^m=\max(d^{\varphi})$ for all values of $\varphi$ in $\rho$. Each row in $\mathcal{D}^t$ is a data sample $\mathbf{x}_i$ from any $X^{\varphi}$, taken at bin $i$, with the corresponding label $y_i$. In case $d^{\varphi}<d^m$, zero padding is applied to edge shots, i.e. $\rho(i,j)$ is set to $0$ for all shots $j$ where $j<s^{\varphi,\Psi}_{b,s}-(d^m–d^{\varphi,\Psi})/2$ and $j>s^{\varphi,\Psi}_{b,e}+(d^m–d^{\varphi,\Psi})/2$. 

The training dataset $\mathcal{D}^t$ is 
2D where $\rho(i,j)$ is a single measurement value (in Radian) for a particular sensor (bin $i$) at a certain time (shot $j$) within the data sample $\mathbf{x}_i$. In this case, the bin $i$ represents the index of the data sample and $j$ is shot index relative to the start shot of $\mathbf{x}_i$. 
Similarly, $\mathcal{D}^v \in \mathbb{R}^{D^v \times d^m}$ is a two dimensional matrix where $D^v$ is a design parameter that determines the number of samples used in the testing phase. Each sample $\mathbf{x}_i$ in $\mathcal{D}^v$ is associated with a label. 
The sum of $(D^t+D^v)\leq (2\times 5B)$, since for each load, the selected car passes by each bin within the controlled stretch of the fibre. Moreover, it passes twice through each bin during the recording of the full dataset~$\rho$; once for $\Psi=east$ and once for $\Psi=west$.

\subsection{HOV or LOV (2-way)}\label{5wayORbinary}
As discussed in the introduction, ITS applications such as HOV lane control are concerned with knowing whether a vehicle has low or high occupancy and less concerned with the actual number of passengers formulated in Section~\ref{Sec:Occ}. As such, in this section we propose a simplification to the original problem of detecting the exact number of passengers. Instead, we formulate an alternative \textit{classification} problem that aims to distinguish between low occupancy (LOV -one or two passengers) or high occupancy (HOV -three or more passengers). It follows that the binary classification (2-way) problem entails a higher number of data samples per class, as shown in Table~\ref{tab:numdatascenarios}. Consequently, the model is less prone to be overfitting compared to the original 5-way model, as will be confirmed in Section~\ref{sec:Results}. 

However, regrouping the data samples into HOV (majority) and LOV (minority) results in data imbalance (Imbalance ratio $IR=1684/2656=0.63$) which may cause a bias in the model. To examine the influence of imbalanced class, we oversample the LOV class samples from the same dataset in order to meet the number of data samples in HOV class. Results of pre-oversampling and oversampling are shown in Table ~\ref{tab:results} with the latter highlighted with $\dagger$.

\subsection{Vehicle size classification}\label{sec:trans}
The data that relates to vehicle types is similar to that of occupancy detection with some differences. Firstly, there 5 different vehicle types numbered between $[1..5]$; these are regrouped into two classes: \textit{Large} and \textit{Small}. To this end, in the problem formulation we replace $\varphi$ with $\omega$ to indicate the car size, where $\omega=\{0,1,2\}$ and $0$ indicates noise, $1$ small cars, and $2$ large cars. Another difference, as seen in Figure~\ref{fig:2Dplot}, is the control vehicles drive in a queue in a single direction. As such, the $\Psi$ is dropped from the problem formulation:  $\mathbf{x}_{i,\Psi} \rightarrow \mathbf{x}_i $. The last difference is that these five vehicles were driven at five different speeds whereby the data from each experiment is collected in a separate dataset. Thus, there are five different datasets $\{\mathbf{X}_{30}\dots \mathbf{X}_{\nu}\dots \mathbf{X}_{70}\}$ where $\nu=\{30,40,50,60,70\}$ is the speed of the vehicles in each experiment. For simplicity of mathematical notation and unless more than one dataset is used at the same time, we use the notation $\mathbf{X}$ to indicate a given dataset without the subscript $\nu$. The vehicle detection problem can thus be formulated as a \textit{classification} predictive modelling; a process of predicting vehicle sizes that are categorised into $\omega>0$ classes/labels (large car and small car.) by approximating a mapping function from input data samples in $\mathbf{X} = \{X^1, X^2\} $ into discrete output labels $Y$. 

\section{Methodology}\label{sec:meth}
In this section, we first describe the method employed to frame events of interest from the raw DAS dataset~$\rho$ and associate a label based on our knowledge of the controlled experiment. Next, we present the CNN structure used for extracting DAS features that delineate the vehicle occupancy and size. \vspace{-0.4cm}
\subsection{Sample labelling}\label{sec:data}
As seen in Figures~\ref{fig1} and~\ref{fig:2Dplot}, there is a linear relationship between the fibre shot $s$ and the fibre bin $b$ for each car trip, as can be excepted given the fixed speed movement. Two linear regressions are applied to frame the start shot $s_{b,s}$ and the end shot $s_{b,e}$ of an event of interest in bin $b$. Let $(c1, c2)$ be coefficients to describe a linear function: $c1\times b+c2 = s_{b,s}$ and the same as $(c3, c4)$ for $c3\times b+c4 = s_{b,e}$. To solve $(c1, c2)$ and $(c3, c4)$, $s_{b,s}$ and $s_{b,e}$ are manually collected in every 50 consecutive bins where $250\leq b\leq 750$. Then we can use the two linear functions to get $s_{b,s}$ and $s_{b,e}$ with any given $b|250\leq b\leq 750$. 

For occupancy detection, each event of interest is labelled as $\varphi = {1, 2, 3, 4, 5}$ which is the number of passengers as mentioned in Section~\label{sec:DAS data}. Therefore, for given $s_{b,s}$ and $s_{e,s}$ shots, each event of interest is a data sample $\mathbf{x}_b = \rho(b,s_{b,s}), \rho(b,s_{b,s}+1), \rho(b,s_{b,s}+2), \rho(b,s_{b,s}+3), \dots, \rho(b,s_{b,e})$.

For vehicle classification, each car window is labelled as either $\omega=0$ if the window contains an unwanted signal, or $\omega=\{1,2\}$ if it contains one of the controlled cars, in which case the value of $\omega$ is determined based on the predefined size of the vehicles (see Section~\ref{sec:dataA}).

We have applied this method to the three different datasets: RC-60-Mix, RC-60-5p, and AllCars-1p described in Section~\ref{sec:dataA}. The results of the sample extraction and labeling are summarised in Table~\ref{tab:numdatascenarios} for occupancy and Table~\ref{tab:classificationdata} for vehicle size.

\begin{table}[] 
\centering
\begin{tabular}{|c|c|c|c|}
\hline
\textbf{Number of }                         & \textbf{Number of }            & \textbf{Occupancy}             & \textbf{Number of}                \\ 
\textbf{passengers}                                & \textbf{data samples}          & \textbf{Low/High}              & \textbf{data samples}             \\ \hline
\multicolumn{1}{|c|}{\textit{RC-60-Mix}} & \multicolumn{1}{l|}{} & \multicolumn{1}{l|}{} & \multicolumn{1}{l|}{}    \\ \hline
1                                           & 782                   & \multirow{2}{*}{LOV}   & \multirow{2}{*}{1684}    \\ \cline{1-2}
2                                           & 902                   &                       &                          \\ \hline
3                                           & 902                   & \multirow{3}{*}{HOV}   & \multirow{3}{*}{2656}    \\ \cline{1-2}
4                                           & 852                   &                       &                          \\ \cline{1-2}
5                                           & 902                   &                       &                          \\ \hline \hline
\multicolumn{1}{|l|}{\textit{Independent Dataset}} & \multicolumn{1}{l|}{} & \multicolumn{1}{l|}{} & \multicolumn{1}{l|}{}    \\ \hline
1 (AllCars-1p)                                           & 451                   & LOV                    & \multicolumn{1}{c|}{451} \\ \hline
5 (RC-60-5p)                                          & 451                   & HOV                    & \multicolumn{1}{c|}{451} \\ \hline
\end{tabular}
\caption{Number of samples in occupancy datasets RC-60-Mix for specific number of passengers and HOV/LOV categorisation. The number of samples the independent dataset is also shown.}\label{tab:numdatascenarios}
\end{table}

\begin{table}[h!]\label{tab:classificationdata}
\centering
{\begin{tabular}{|l|l|l|l|l|l|}
\hline
Car Size\textbackslash{}Speed (km/h) & 30   & 40   & 50   & 60   & 70   \\ \hline
Large                           & 501  & 803  & 918  & 902  & 1102 \\ \hline
Small                           & 1503 & 1154 & 1377 & 1353 & 1653 \\ \hline
\end{tabular}%
}
\caption{Number of samples for each dataset in AllCars-1p where each dataset corresponds to a different fixed car speed.}
\end{table}

\subsection{Feature extraction and classification}\label{sec:classification}
In this section, we design a CNN network to extract the key features that would allow for the best estimation of car size and occupancy. We first examine the efficacy of a 1D-CNN, as detailed in Section~\ref{sec:1D}. Next, we present a novel two-dimensional CNN (2D-CNN) with the aim of capturing more descriptive features of the weight distribution that would result from the car occupancy, as discussed in Section~\ref{sec:2D}.
\begin{figure*}[t!]
	\centering
	\includegraphics[width=0.5\textwidth]{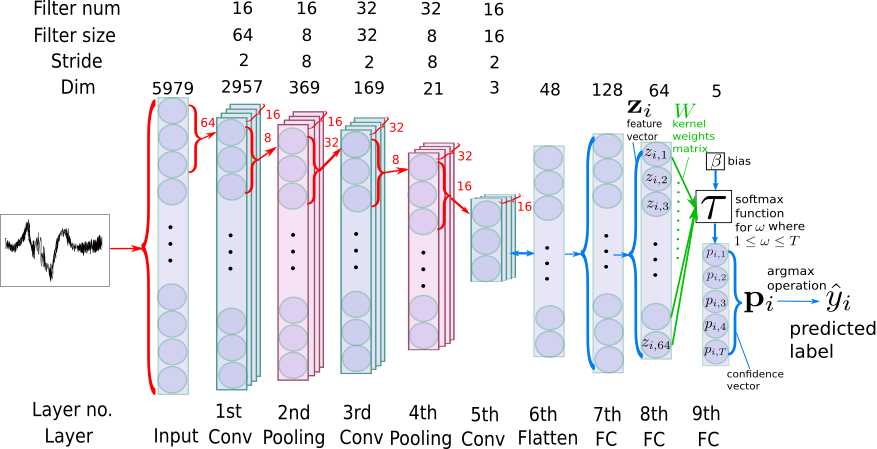}
	\caption{In this figure we show proposed 1D-CNN architecture for both 5-way occupancy detection and vehicle classification. For the former, the softmax function is replaced with an SVM classifier.}
	\label{fig:1DCNN}
\end{figure*}

\begin{figure*}[t!]
	\centering
	\includegraphics[width=0.75\textwidth]{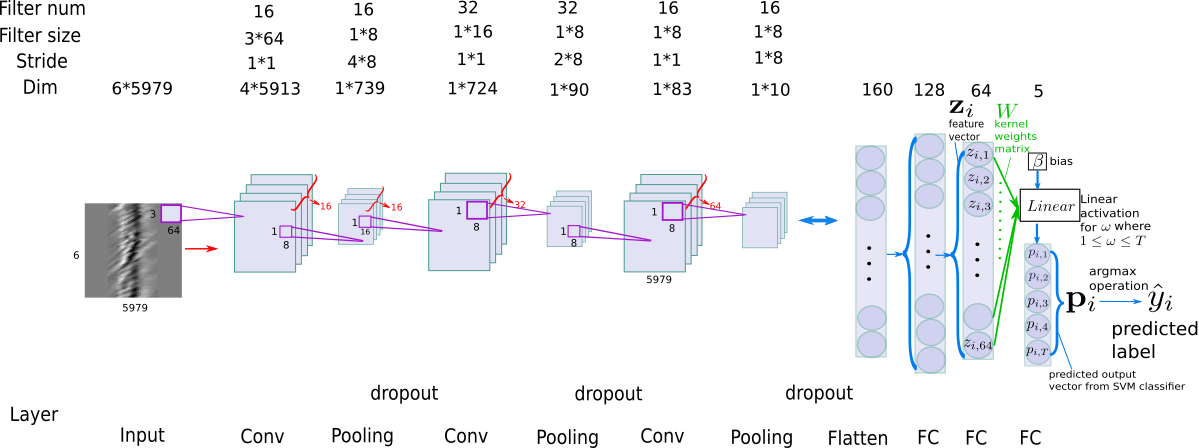}
	\caption{The 2D-CNN architecture is used for vehicle occupancy detection. It is similar to 1D, however, since the first dimension was small, we limited the width of convolution stride size (set to 1) to preserve more information in the first dimension. The parameters of Flatten, Fully Connected, and Classification layers have the same setting as 1D-CNN.}
	\label{fig:2DCNN}
\end{figure*}

\subsubsection{1D-CNN for feature extraction}\label{sec:1D}
We first adopt a 1D-CNN to extract the sample features followed by a Support Vector Machine (SVM) classifier {\color{black} in the case of occupancy detection and a softmax classifier for vehicle size classification.} 
In occupancy detection, for each input signal $\mathbf{x}_i$ with index $1\leq i\leq B$, the classifier produces an output $\mathbf{p}_i=\{p_{i,1},.., p_{i,\varphi},..,p_{i,T}\}$ such that $\mathbf{p}_i$ represents a classification result and will be sent to an argmax operation to define the final predicted label. 
The 1D-CNN is composed of fourteen layers, as shown in Figure~\ref{fig:1DCNN}. The output of the eighth layer (Fully connected) is a features' vector $\mathbf{z}_i=\{z_{i,1}, \dots,z_{i,Z}\}$ that corresponds to each input sample $\mathbf{x}_i$ and where $Z=64$. This features' vector $\mathbf{z}_i$ is thus the input of a predefined classifier. 

Consider a vector $\mathbf{k}_i=\mathbf{W} \mathbf{z}_i + \boldsymbol{\beta}$, where $\mathbf{W}$ represents the weights of the CNN network, $\mathbf{z}_i$ is the feature vector, and $\boldsymbol{\beta}=\{\beta_1, \beta_2, \cdots, \beta_5\}$ represents the set of bias values associated with each class. Both $\mathbf{W}$ and $\boldsymbol{\beta}$ are trained and updated through the BP process. The SVM  linear activation function, each vector $\mathbf{k}_i = \mathbf{p}_i$, since $\mathbf{p}_i$ is predicted output vector contains individual predicted value $p_i$; therefore, $\mathbf{p}_i = [p_{i,1}, p_{i,2}, p_{i,3}, \dots, p_{i,5}]$ for all classes $\varphi = {1,2,3, \dots, 5}$. For each $p_{i,\varphi}$, $||p_{i,\varphi}||$ geometrically represents the distance between given data point $\mathbf{x}_i$ and the hyperplane that decides whether $\mathbf{x}_i$ is classified as $\varphi$. In this case, $\hat{y}_i= \mathop{\mathrm{argmax}}(\mathbf{p}_i)=\varphi$.

An SVM classifier is selected as it is known to work well when data clusters have clear margins. During the model training, kernels are optimised through backpropagation (BP). BP calculates a partial derivative of a loss function; for SVM the loss function the sum of Huber Loss shown in Eq(\ref{eq:Hub}).
\begin{equation}
\mathrm{H}(i)
= 
\left\{ \begin{matrix} \frac{1}{2} a_i^2 & \text{if } |a_i| < \delta \\
\delta (|a_i| - \frac{1}{2}\delta ) & \text{otherwise} \end{matrix} \right.
\label{eq:Hub}
\end{equation}

Where $a_i = \mathbf{p}_i - \mathbf{g}_i$ and $\delta$ is a threshold which sets to constant $1$. $\mathbf{g}_i$ is ground truth vector which, for each element $g_{i,\varphi}$ is equal to $1$ if $\varphi=y_i$ else $g_{i,\varphi}$ is equal to $0$. In this case, $\mathbf{g}_i = [g_{i,1}, g_{i,2}, g_{i,3}, \dots ,g_{i,\varphi}, \dots, g_{i,5}] = [0, 0, 0, \dots, 1, \dots, 0]$.

While in vehicle size classification SVM is replaced by softmax classifier, for softmax the loss function is based on the cross-entropy $\mathrm{CE}(i)=-\log(p_{i,\omega})$ or negative log likelihood. In this case, $\omega=y_i$ is the ground truth label of sample $\mathbf{x}_i$ and $p_{i,\omega}$ is the confidence level that is produced by the softmax function $\tau(\mathbf{u}_i) = \mathbf{p_{i}}$ which transforms the vector $\mathbf{u}_i$ to a vector of confidence levels $\mathbf{p_{i}} = [p_{i, 1}, p_{i, 2}, \dots, p_{i, T}]$ for each input data sample $\mathbf{x}_i$, where $p_{i,\omega} = \frac{e^{u_{i,\omega}}}{\sum_{j=1}^{T} e^{u_{i,j}}}$.

The CNN updates weights in the kernel such that $H(i)$ or $\mathrm{CE}(i)$ is decreased after each batch of 32 input data samples $x_i$ is processed. The goal of model training is to minimise the total loss $L$ which is sum of individual $H(i)$ (or $\mathrm{CE}(i)$) for all samples from the training set $\mathcal{D}^t \in \mathbb{R}^{D^t \times d^m}$  and the testing set $\mathcal{D}^v \in \mathbb{R}^{D^v \times d^m}$ .

After passing input to convolution layer, the extracted features are compressed by a maxpooling layer to exclude most feature information but retain the max pixel value in each pre-defined region. It is said to preserve important features and save computing power. 
Differently from the network shown in Figure~\ref{fig:1DCNN}, a Dropout layer is added after maxpooling layer when applied to the occupancy detection problem. The dropout layer randomly set number of input values to 0 so the neurons in current layer are not activated. By doing so, CNN is forced to train itself with limited feature information and is less likely to be overfitting; a known risk in occupancy detection due to the limited number of data samples.   

\subsubsection{2D-CNN for feature extraction}\label{sec:2D}
The 1D DAS signal used in Section~\ref{sec:1D} does not include spatial information about moving objects. We presume that such spacial information contains features that would indicate the vehicle occupancy, since this is related to weight distribution over the area of a vehicle. To this end, we posit here that a 2D CNN can capture spatial features, hence, provide better performance than a 1D CNN for this given problem. 

To generate 2D spatial-temporal signals, we used the sliding window method with window size of six bins (each bin covers $0.68$ meters), which is $6 \times 0.68=4.08$ metres (the Renault Clio is $4.05$~metres in length), and stride was one bin. As shown in Figure~\ref{fig:2DCNN}, we add dropout layers to mitigate the overfitting problem, as in  Section~\ref{sec:1D}. 

\vspace{-0.3cm} 

\section{Experimental results and analysis}\label{sec:Results}
In this section, we present the results and interpretation for the three defined problems in Sections~\ref{Sec:Occ},~\ref{5wayORbinary}, and~\ref{sec:trans}.
\subsection{Vehicle occupancy detection (5-way and 2-way)}
\begin{table}[t!]
\centering
\begin{tabular}{|ll|llllll|}
\hline
\multicolumn{2}{|l|}{Problem}                               & \multicolumn{2}{c|}{5-way}                            & \multicolumn{2}{c|}{2-way}                          & \multicolumn{2}{c|}{2-way†}    \\ \hline
\multicolumn{2}{|l|}{Model}                                 & \multicolumn{1}{l|}{1D}    & \multicolumn{1}{l|}{2D}  & \multicolumn{1}{l|}{1D}  & \multicolumn{1}{l|}{2D}  & \multicolumn{1}{l|}{1D}  & 2D  \\ \hline
\multicolumn{2}{|l|}{Nb of Epochs}                          & \multicolumn{1}{l|}{1000}  & \multicolumn{1}{l|}{500} & \multicolumn{1}{l|}{100} & \multicolumn{1}{l|}{500} & \multicolumn{1}{l|}{100} & 500 \\ \hline
\multicolumn{2}{|l|}{Train:Test}                            & \multicolumn{1}{l|}{67:33} & \multicolumn{5}{c|}{80:20}                                                                                      \\ \hline
\multicolumn{1}{|c|}{\multirow{4}{*}{Acc (\%)}} & Main RC-60-Mix      & \multicolumn{1}{l|}{66}    & \multicolumn{1}{l|}{92}  & \multicolumn{1}{l|}{85}  & \multicolumn{1}{l|}{97}  & \multicolumn{1}{l|}{87}  & 97  \\ \cline{2-8} 
\multicolumn{1}{|c|}{}                          & Ind. AllCars-1p & \multicolumn{1}{l|}{45}    & \multicolumn{1}{l|}{69}  & \multicolumn{1}{l|}{56}  & \multicolumn{1}{l|}{50}  & \multicolumn{1}{l|}{60}  & 53  \\ \cline{2-8} 
\multicolumn{1}{|c|}{}                          & Ind. RC-60-5p & \multicolumn{1}{l|}{1}     & \multicolumn{1}{l|}{6}   & \multicolumn{1}{l|}{69}  & \multicolumn{1}{l|}{77}  & \multicolumn{1}{l|}{56}  & 75  \\ \cline{2-8} 
\multicolumn{1}{|c|}{}                          & Ind. Avg  & \multicolumn{1}{l|}{23}    & \multicolumn{1}{l|}{38}  & \multicolumn{1}{l|}{63}  & \multicolumn{1}{l|}{64}  & \multicolumn{1}{l|}{58}  & 64  \\ \hline
\end{tabular}
\caption{Summary of results highlighting the performance of both models for three occupancy problems 5-way, 2-way and 2-way$\dagger$ (after over-sampling to balance the dataset). The accuracy from main an independent (Ind.) dataset is shown.}\label{tab:results}
\end{table}
Both 1D-CNN and 2D-CNN models were trained to solve both occupancy detection problems (5-way and 2-way) based on the RC-60-Mix dataset with training/testing ratio as indicated in Table~\ref{tab:results}. Another set of results in shown for the 2-way problem in which oversampling is applied to the minority class (LOV) before training the model to compensate for the data imbalance; this is denoted as 2-way$\dagger$ in Table~\ref{tab:results}. For each set of results, the average accuracy of occupancy detection is presented, based on the main dataset RC-60-Mix. For further validation, an independent dataset was patched using RC-60-5p to represent five passengers (HOV in the 2-way problem) and the RC-related samples at 60km/h in the AllCars-1p dataset were used to represent one passenger (LOV in the 2-way problem). 

There are a few insights that can be drawn by examining these results, related to: model validity/overfitting, 1D versus 2D representation, impact of data imbalance, and computational cost. The overarching insight is that the hypothesis that the DAS signal contains information about the number of passengers in a known car is confirmed. It is indeed possible to detect the exact number of passengers with $92\%$ and that of the binary occupancy by $97\%$ with the 2D-CNN model. However, the model suffers from overfitting, which is apparent when the independent average accuracy is calculated. As seen in Table~\ref{tab:results}, the performance of these models drops to $38\%$ and $64\%$, respectively. Nonetheless, the results are promising and indicate that it should be possible to avoid overfitting if it were possible to obtain larger datasets. 

As suspected and discussed in Section~\ref{sec:classification}, the information concerning the car occupancy is better captured through 2D DAS samples as opposed to 1D samples. This is due to the fact that the information relates to the weight distribution over the area of the car as opposed to absolute weight figure. This is confirmed by examining the results in Table~\ref{tab:results} in which the 2D-CNN results systematically outperform the 1D-CNN in each of the three scenarios: 5-way, 2-way, and 2-way$\dagger$.

The effect of the data imbalance in the case of 2-way classification is evident when examining the corresponding results. With an imbalance ratio of $IR=0.63$, the 2-way models for both 1D-CNN and 2D-CNN are clearly biased to the majority class: HOV. This is reflected in the HOV independent accuracy (RC-60-5p)  $69\%$ (1D-CNN) and $77\%$ (2D-CNN), in comparison with $56\%$ and $50\%$ for the minority LOV class (AllCars-1p). After oversampling the LOV class $IR=1$, the accuracy for (AllCars-1p) improves for both CNN models, which reconfirms the 2-way models' bias.

Moreover, we can draw on the computational complexity of the model training by examining the number of training epochs required for each model to converge. The 2D-CNN systematically requires more time to converge in comparison with the 1D-CNN. This is expected as the number of tune-able parameters is larger. The only exception to this trend occurs during the training of 5-way models; it may be the case that the 1D-CNN does not converge due to the limited training dataset and limited occupancy information in 1D samples. In order to confirm our analysis, we retrained the 1D-CNN for 5-way classification without the dropout layers after each maxpooling layer. The aim is to preserve any useful information while training at the cost of increasing the number of tune-able parameters. With this setting, the main data accuracy reached $87\%$ after $100$ epochs. We confirm two insights with these results. First, we confirm that the 1D-CNN model is generally less computationally expensive than the 2D-CNN. Second, the 1D samples contain limited information about the vehicle occupancy and do not afford dropping any information in the training process. It follows, that the 1D-CNN for 5-way classification is bound to be overfitting unless more data is available for training. 

\subsection{Vehicle size detection}
\begin{table}[]
\centering
{%
\begin{tabular}{|l|l|}
\hline
Problem      & vehicle size classification     \\ \hline
Model        & 1D                              \\ \hline
Nb of Epochs & 100                             \\ \hline
Train:Test   & 67:33                           \\ \hline
Main Acc.\% (AllCars-1p) & Avg 92 (Large: 89, Small: 94) \\ \hline
Ind. Acc.\% (RC-60-Mix) & Avg 72 (1p:76, 2p:83, 3p:66, 4p:67, 5p:66)\\ \hline
\end{tabular}\caption{Summary of results of vehicle size classification with main and independent datasets.}\label{tab:results2}
}
\end{table}
The results of the third problem, vehicle size classification, are summarised in Table~\ref{tab:results2}. In this case, the 1D-CNN model was trained based on 67\% of the AllCars-1p dataset which includes all of the five experiments described in Section~\ref{sec:sizeData}. Since each of these experiments relates to the same five vehicles driven with a single passenger but at different speeds, the aim of mixing samples is to discern features that represent the size of the vehicle despite the often dominant speed characteristics. The results of the model are first tested based on 33\% of AllCars-1p, as shown in Table~\ref{tab:results2}. Next, we validate further model by testing is on the RC-60-Mix samples to examine if the number of passengers affects the vehicle size classification. In this case, the aim is to separate vehicle size features from the absolute weight affected by the number of passengers. The average difference in weight between Small and Large vehicles is 350~Kg and the average weight of one passenger is 80~Kg; thus, four additional passengers weigh 320~Kg which renders the total weight of the RC comparable to that of large vehicles.

The main accuracy (AllCars-1p) results clearly confirm that the features delineating the type of vehicle can be successfully extracted with 1D-CNN despite the dominating speed feature. The results demonstrate that a 1D DAS data representation is sufficient to capture the salient features that determine the vehicle size as can be deduced from the promising accuracy results both in large car: 89\% and small car: 94\% (see Table~\ref{tab:results2}). 

In order to validate the findings further, we use samples from RC-60-Mix as independent validation dataset. In this case, the same Renault Clio is driven along the same stretch of the road carrying different number of passengers between one and five. As shown in Table~\ref{tab:results2}, the model is able to correctly classify the size of the Renault Clio $72\%$ for any occupancy. This finding suggests that the large and small vehicle class inherits distinct information that is not overshadowed by the speed nor the occupancy of the vehicle. However, it is clear that higher occupancy cause higher error in detecting the size of the vehicle which suggests that a 2D-CNN, in this case, would yield better data representation.
\section{Conclusion}\label{sec:con}
In this manuscript, we present a pioneering research that uses signals generated by a Distributed Acoustic Sensor (DAS) system for intelligent transportation. We formulate two dominant problems in this domain: occupancy estimation of moving vehicles and vehicle size classification. Based on real DAS dataset collected in a controlled experiment, we demonstrate that the DAS signal generated by a moving car is indicative of its size and occupancy level. We propose two different convolutional neural networks (CNN) to extract the spacio-temporal features of the DAS signal which successfully classifies the level of occupancy (HOV/LOV) with $97\%$ accuracy and estimates the number of passengers with $92\%$ accuracy. When the models are further validated using an unseen independent dataset, the accuracy figures drop to $38\%$ and $64\%$, respectively, highlighting a model overfitting problem. In the case of vehicle size classification, the 1D model succeeds in determining the size of a car irrelevant for the speed of movement with 92\% accuracy. With an independent test, we further demonstrate the model in not affected by the occupancy of the vehicle as it is still capable of determining the size 72\% of the time.
\bibliographystyle{ieeetr}
\bibliography{bibtex/bib/DASpassenger.bib}



\end{document}